\documentclass[sigconf, review=false]{acmart}
\pdfoutput=1

\usepackage{libertine}
\usepackage{booktabs} 
\usepackage{times}
\usepackage{epsfig}
\usepackage{graphicx}
\usepackage{amsmath}
\usepackage{mathtools}
\usepackage{multirow}
\usepackage{multicol}
\usepackage{caption}
\usepackage{subcaption}
\usepackage{adjustbox}
\usepackage{enumitem}
\usepackage{colortbl}
\usepackage{natbib}
\usepackage{listings}
\usepackage{xcolor}
\usepackage{ragged2e}
\usepackage{blindtext}

\setcopyright{acmlicensed}

\acmDOI{10.475/123_4}

\acmISBN{123-4567-24-567/08/06}

\acmConference[ICVGIP'22]{13th Indian Conference on Computer Vision, Graphics and Image Processing}{December 2022}{Gandhinagar, India}
\acmYear{2022}
\copyrightyear{2022}
\title{ContextCLIP: Contextual Alignment of Image-Text pairs on CLIP visual representations}
\author{Chanda Grover}
\affiliation{
\institution{Ashoka University}
\city{Sonipat, Haryana}
\country{India}
}
\author{Indra Deep Mastan}
\affiliation{
\institution{LNMIT}
\city{Jaipur, Rajasthan}
\country{India}
}

\author{Debayan Gupta}
\affiliation{
\institution{Ashoka University}
\city{Sonipat, Haryana}
\country{India}
}
\begin{abstract}
State-of-the-art empirical work has shown that visual representations learned by deep neural networks are robust in nature and capable of performing classification tasks on diverse datasets. For example, CLIP demonstrated zero-shot transfer performance on multiple datasets for classification tasks in a joint embedding space of image and text pairs. However, it showed negative transfer performance on standard datasets, e.g., BirdsNAP, RESISC45, and MNIST. In this paper, we propose ContextCLIP, a contextual and contrastive learning framework for the contextual alignment of image-text pairs by learning robust visual representations on Conceptual Captions dataset.
Our framework was observed to improve the image-text alignment by aligning text and image representations contextually in the joint embedding space. ContextCLIP showed good qualitative performance for text-to-image retrieval tasks and enhanced classification accuracy. We evaluated our model quantitatively with zero-shot transfer and fine-tuning experiments on CIFAR-10, CIFAR-100, Birdsnap, RESISC45, and MNIST datasets for classification task. 
\end{abstract}

\begin{document}
\maketitle
\titlenote{}
%
%

\section{Introduction} 
\begin{figure}
\begin{multicols}{2}
    \includegraphics[width=0.93\linewidth]{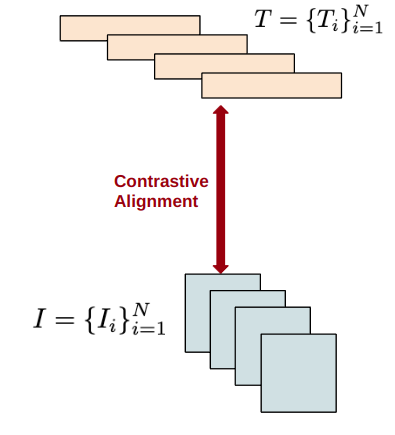}\par
    \subcaption{\small Image-Text Alignment through CLIP \cite{radford2021learning}}
    \label{fig:problem_defa}
    \includegraphics[width=0.98\linewidth]{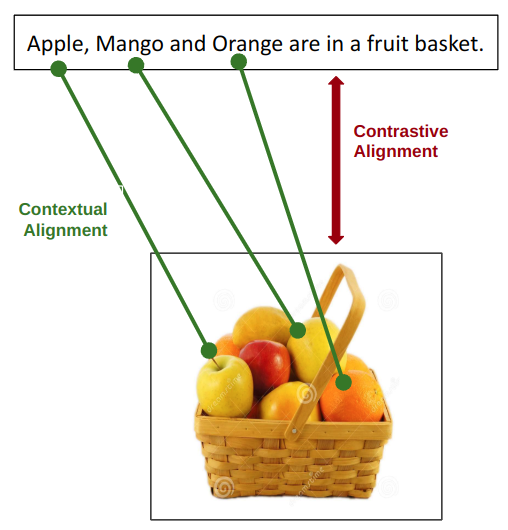}\par
    \subcaption{\small Image-Text Contextual Alignment (ContextCLIP)}
    \label{fig: problem_defb}
\end{multicols}
    \caption{The figure (a) illustrate the contrastive alignment of the image-text pairs as performed by CLIP \cite{radford2021learning}. (b) shows the contextual alignment of the image-text pair as performed by ContextCLIP.}
    \label{fig:problem_def}
\end{figure}

Learning robust image representations is an enduring problem in computer vision, as is the related question of making those representations transferable to some other dataset. Image representations also contain linguistic context, making a transfer to other domains a reasonable problem to pursue. No other benchmarks have been pursued as religiously as the ImageNet \cite{deng2009imagenet, russakovsky2015imagenet} network architectures for a broad array of computer vision problems like transferring to new datasets \cite{donahue2014decaf, sharif2014cnn}, object detection \cite{huang2017speed}, image segmentation \cite{chen2017deeplab} etc. Recently, natural language-based supervision has been extensively researched for learning robust image representations, e.g., CLIP \cite{radford2021learning}, ALIGN \cite{jia2021scaling}, and BASIC \cite{pham2021combined} have enjoyed great success in extending contrastive learning to paired image-text data, with impressive zero-shot classification and robustness. Visual self-supervision \cite{mu2021slip} and use of unpaired data \cite{Tejankar2021fistful} have been included in achieving such robust visual representations. Although these methods have achieved good performance in terms of positive zero-shot transfer on multiple datasets, they have shown negative zero-shot transfer on some datasets like MNIST, RESISC45, and BirdsNap. 
\\~
Contrastive learning only exploits the high-level image-text pair similarity information but does not exploit semantic-level image-text pair similarity; it does not capture the contextual alignment of image and text pairs. Our work identifies contextual learning complements the contrastive learning objective for image-text alignment. 
Fig.~\ref{fig:problem_def} depicts the addition of contextual information to image-text pairs. As shown in Fig.~ \ref{fig:problem_defa}, CLIP enables the alignment of image-text pairs through contrastive learning. It only aligns $I_1$ with $T_1$ and not with $T_2$ and so on. Somewhat more formally, the image-text alignment problem can be described as follows: given a set of images and captions, determine the relation between image-text pairs based on the \textit{contextual similarity}. The standard contrastive learning objective aims to identify the matched image text pairs (''positives'') against the mismatched (''negatives'') image-text pairs. \cite{chen2020simple,oord2018representation}. This, however, does not make the alignment of $I_1$ and $T_1$ at the semantic level. For example,  Fig.~\ref{fig: problem_defb} illustrates words such as orange and mango of captions $T_1$ are matched with regions corresponding to mango and orange in image $I_1$.

In this paper, we aim to learn robust image representation by studying image-text contextual alignment, 
i.e., given valid image-text pairs, we compute the image and text embedding that can be contextually aligned in the embedding space. Such an objective can be fulfilled when one can align the learned image and text embeddings at the context level. We hypothesize that because of the contextual non-alignment of the embeddings (of true image-text pairs), CLIP showed negative zero-shot transfer on standard datasets, e.g., BirdsNAP, RESISC45, and MNIST. To fix this, we propose ContextCLIP, a framework for the contextual alignment of image-text pairs based on the robust visual representation of CLIP using contextual loss \cite{mechrez2018contextual}. Contextual loss aligns image-text pairs at the semantic level; specifically, given an image-text pair learned through contrastive learning, we augment the contrastive learning objective with a contextual learning objective at the feature level. The latter uses the nearest neighbor field to measure the similarity between images and texts. It considers a particular feature of one modality and identifies the ``most similar'' of the other. These features are matched in a projected space of image and text modalities (low-dimensional) and introduces contextual alignment at the feature level by reducing the distance between the image and text embeddings.
We show that explicit formulation of contextual alignment objective improves the overall alignment between image-text pairs, as shown in Tables \ref{tab:zeroshot} and \ref{tab:fine_tuning}.
In all cases, we pre-train our models on a small fraction of the Conceptual Captions 3M dataset \cite{sharma2018conceptual} (around 6K image-text pairs) and evaluate them on CIFAR-10, CIFAR-100, MNIST, RESISC45, BirdsNap with zero-shot transfer and fine-tuning experiments on 
\twocolumn[{%
\renewcommand\twocolumn[1][]{#1}%
\begin{center}
    \centering
    \begin{minipage}{0.16\linewidth}\centering
	Text Description 
    \end{minipage}
    \begin{minipage}{0.41\linewidth}\centering
	(a) CLIP \cite{radford2021learning}
    \end{minipage}
     \hfill
    \begin{minipage}{0.41\linewidth}\centering
	(b) ContextCLIP  (ours)
    \end{minipage}
    \begin{minipage}{0.16\linewidth}
    A little girl climbing on red roping
    \end{minipage}
    \begin{minipage}{0.41\linewidth}
        \includegraphics[width=0.98\linewidth]{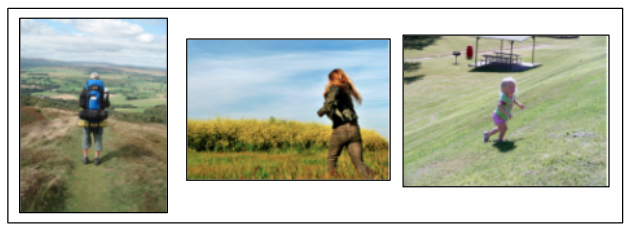}
    \end{minipage}
     \hfill
    \begin{minipage}{0.41\linewidth}
        \includegraphics[width=0.98\linewidth]{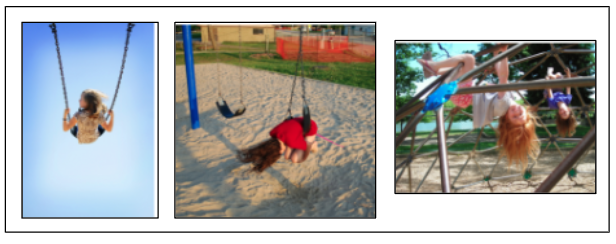}
    \end{minipage}
    \begin{minipage}{0.16\linewidth}
    A small boy putting something in his mouth with both hands
    \end{minipage}
    \begin{minipage}{0.41\linewidth}
        \includegraphics[width=0.98\linewidth]{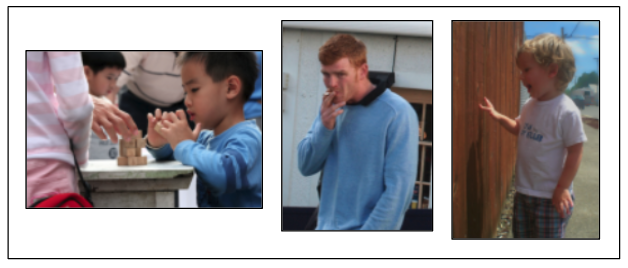}
    \end{minipage}
    \begin{minipage}{0.41\linewidth}
        \includegraphics[width=0.98\linewidth]{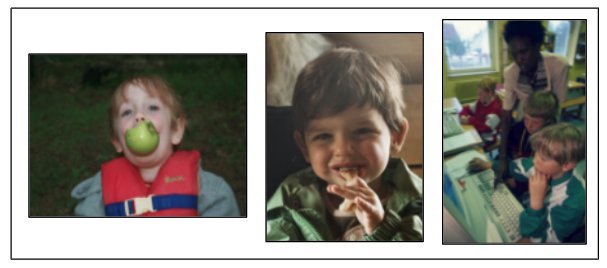}
    \end{minipage}
\end{center}%
    \captionsetup{type=figure}
    \captionof{figure}{The figure shows the output comparison between (a) CLIP \cite{radford2021learning} and (b) ContextCLIP (ours) for text-to-image retrieval task.\\~\\~}
    \label{fig:title}
}]
standard datasets. For qualitative performance evaluation, we performed text-to-image retrieval experiments; Fig.~\ref{fig:title} and Fig.~\ref{fig:t2ir} show text-based image retrieval results for CLIP and ContextCLIP. The results of our proposed framework are visually better in terms of consistency with the textual descriptions. 
\\~
\\~
\noindent\textbf{Contributions.} The contributions are summarized as follows.
\begin{itemize}[leftmargin=*]
\item We analyze contrastive learning for representation learning jointly over image and text modalities and identify non-alignment between the representation space of image and text modalities of CLIP (Fig.~\ref{fig:ie} and Fig.~\ref{fig:te}). 
\item We propose ContextCLIP, a framework for contextual alignment of image-text space by employing contextual loss between non-aligned feature space of text and image embeddings (Sec.~\ref{sec:approach}).
\item We show that our approach learns a more robust visual representation than CLIP and shows positive zero-shot transfer on standard datasets where CLIP had negative zero-shot transfer performance. To show better representations obtained, we also performed fine-tuning experiments. (Table \ref{tab:zeroshot} and \ref{tab:fine_tuning}).
\end{itemize}

\begin{figure*}[!htb]
    \centering
    \includegraphics[width=0.98\textwidth]{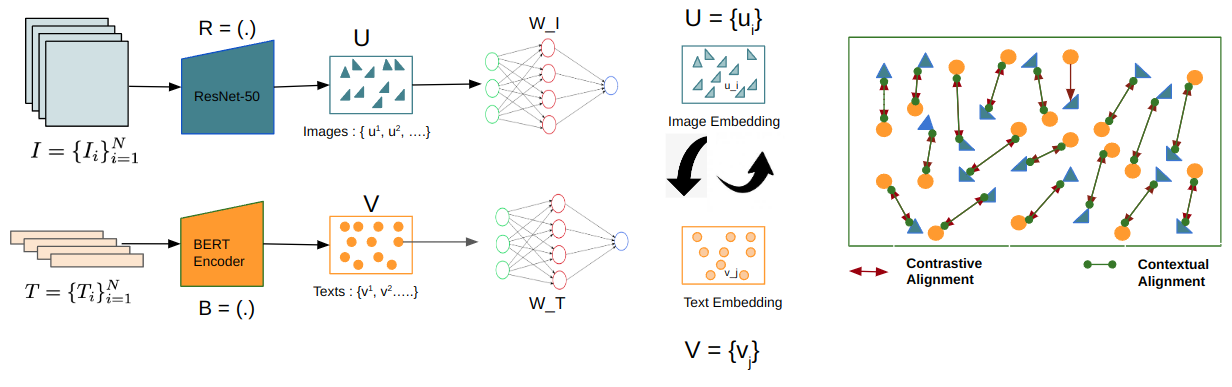}
    \caption{The figure shows the major components of ContextCLIP framework. It consists of image and text encoders with contextual and contrastive alignment objectives. The key idea is that contextual and contrastive alignment at embedding space helps to perform the image-Text alignment. The maroon arrow between the filled orange-colored circle (Text) and blue-colored triangle (Image) shows the contrastive alignment. The green arrow indicates the contextual alignment between Image-Text Pair.}
    \label{fig:contrastive_contextual}
\end{figure*}

\begin{figure*}[!htb]
    \centering
    \begin{subfigure}{0.18\linewidth}
    \includegraphics[width=0.98\linewidth]{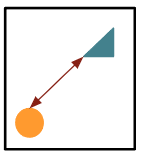}
    \caption{}
    \label{fig:llda}
    \end{subfigure}
    \hfill
    \begin{subfigure}{0.30\linewidth}
    \includegraphics[width=0.98\linewidth]{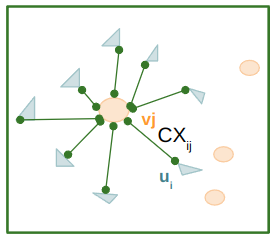}
    \caption{}
    \label{fig:lldb}
    \end{subfigure}
    \hfill
    \begin{subfigure}{0.34\linewidth}
    \includegraphics[width=0.98\linewidth]{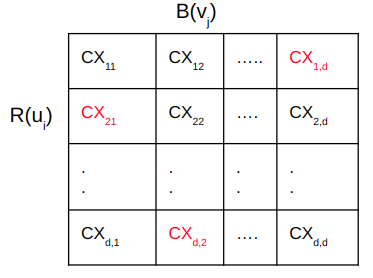}
    \caption{}
    \label{fig:lldc}
    \end{subfigure}
    \hfill
    \begin{subfigure}{0.14\linewidth}
    \includegraphics[width=0.98\linewidth]{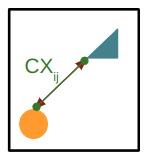}
    \caption{}
    \label{fig:lldd1}
    \end{subfigure}
    \caption{The figure shows contextual learning of Image-Text pairs. (a) Contrastive alignment of image (triangle) and text pair (circle) obtained through CLIP \cite{radford2021learning, Yuan2021FlorenceAN, sun2021lightningdot} (b) Many-to-One correspondence between text feature points (light orange colored circle $v_j$ ) and image feature points (light cyan colored triangle $u_i$). $CX_{ij}$ is the contextual similarity between $u_i$ and $v_j$. (c) Affinity matrix of contextual similarity between $u_i$ and $v_j$. In each column, highlighted red color entry represents embedding $u_i$ with the highest contextual similarity to the given $v_j$. (d) shows $u_i$ and $v_j$ are aligned contextually in addition to the contrastive alignment.}
    \label{fig:LLD}
\end{figure*}

\section{Related Work}\label{sec:rlwork}
Multimodal data is learned in a way that can represent and summarize the complementary information of all modalities and remove redundant information. It can be achieved either through joint representations or coordinated representations. Collective representations have been used in state-of-the-art vision-language pre-training models \cite{coca2022yu}. Further, to identify the direct relations between sub(elements) from two or more different modalities, two types of alignments have been used i.e., implicit and explicit. We have used both types of alignment, implicit (i.e., contrastive) as well as explicit (i.e., contextual), to find the relation between sub(elements) of two modalities.   
\\~
\\~
\noindent \textbf{Contrastive representation learning.} It was originally proposed for self-supervised representation learning in the unimodal context where the embeddings of a sample are brought closer to an augmented version of the sample. In contrast, the embeddings are pushed away for other samples, and their augmentations \cite{gutmann2010noise} \cite{chen2020simple, gao-etal-2021-simcse, oord2018representation, he2020momentum, grill2020bootstrap}. Some works like \cite{zbontar2021barlow}  impose additional constraints to remove redundancies and prevent dimensional collapse in the visual representations. Contrastive learning has also been used to learn robust representations of the multimodal data \cite{Yuan2021CVPR, pmlr-v139-ramesh21a}.

The image and text multimodal data is encoded separately with their respective encoders, producing individual vision and language vectors embedded into a joint space using a contrastive loss. Contrastive approaches have the capability to learn a highly generic visual representation and have good strength to do classification on downstream tasks. Moreover, when pre-trained on a large and diverse dataset \cite{jia2021scaling, pham2021combined} strong zero-shot vision-text retrieval and classification performance have been obtained. Unfortunately, as they are only trained to match visual data to text description, these models can only be adapted to close-ended tasks. Finally, it is challenging to adapt contrastive models using a handful of examples. In fact, Radford et al. (2021) indicated that using as few as two training examples per class actually decreased the CLIP zero-shot performance. But, Flamingo models have significantly improved with as few as four examples. Many works use additional losses to imbibe extra supervisory multimodal knowledge during the training process \cite{Singh2022CVPR, zhang2021vinvl, desai2021virtex, mai2022hybrid}. In this work, we focus on having contextual loss\cite{mechrez2018contextual} in addition to the contrastive loss to learn more robust and aligned image-text representations.
\\~
\\~
\noindent \textbf{Contextual Learning.} Contextual learning is useful for various  applications such as  image-to-image transformation tasks  \cite{mechrez2018contextual}, style transfer \cite{mastan2021deepobjstyle, deep2020deepobjstyle} and image restoration \cite{mastan2020dcil, mastan2022dilie}. Mastan et al. have shown contextual learning is also useful for image enhancement when training with limited training samples  \cite{mastan2019multi, deep2019multi, mastan2021deepcfl, deep2020deepcfl}. 
 The key challenge in image-to-image transformation tasks comes in the presence of non-aligned image data, where employing a pixel-based loss function will be less effective. Contextual learning on images first extracts contextual features using a pre-trained feature extractor and then compares the extracted representations using the contextual loss. The critical observation here is that similar features will be closer in the embedding space. Therefore, the contextual loss will be able to match the feature even in the presence of the non-alignment of the image data. We employ contextual loss on the pre-trained embeddings of the image and text pair captured from the feature extractor. 
\\~

\noindent \textbf{Image-Text alignment Models.}
In recent years rapid progress has been made in vision-language pre-training (VLP), which aims to jointly encode vision and language in a fusion model. Image-Text alignment models subsume the vision-language pre-training. CLIP \cite{radford2021learning} and ALIGN \cite{jia2021scaling} demonstrate that dual encoder models pre-trained with contrastive objectives on image-text pairs can learn the strong image and text representations for cross-modal alignment task and zero-shot classification. Contrastive dual approaches \cite{Jing2018CascadeAN} rely on a relatively similar procedure with the steps as follows i) Extract discriminative image features using a deep neural network ii) Extract text features using another deep neural network iii) use a loss function that measures as accurately as possible the distance between two embedding.  These models align image-text pairs only at the data point level but not at the level of semantics of image and text. In this paper, we align the context of image and text by incorporating contextual loss at the joint image-text embedding space. 
\section{Our Approach}\label{sec:approach}
We propose ContextCLIP framework for the alignment of contextual features extracted from image-text pair to enhance the alignment between image and text pair. The overview of our proposed framework is illustrated in Fig.~\ref{fig:contrastive_contextual}. We start with feeding an image and its corresponding text description to image encoder and text encoder respectively to get their encoded representations. These joint image-text representations are then projected to low-dimensional space through the projection head. There, we represent each image and text having the same size subspace. We then measure the similarity between the image and text pairs as a similarity measure between these point sets. Contextual loss is computed at this point, which makes contrastively aligned image-text pairs to be aligned at the feature level. 
\subsection{Model Architecture}
Our models use the same architecture as the original CLIP model presented in \cite{radford2021learning}. In addition to CLIP architecture, we have introduced a projection head for image ($W_I$) after image and text ($W_T$) as shown in Fig.~\ref{fig:contrastive_contextual} to bring different dimensional representational sizes of image and text to be of the same sizes.
\\~
\\~
\noindent \textbf{Image Encoder.} We used ResNet-50 architecture \cite{he2016deep} as the image encoder (R) for learning image representations as shown in Fig.~\ref{fig:contrastive_contextual}. We considered image representations of 2048 size. All images are resized to $224 \times 224$. 
\\~
\noindent \textbf{Text Encoder.} It learns the feature representations from natural language description, which is input to the text encoder as shown in Fig.~\ref{fig:contrastive_contextual}. For the text encoder (B), we used the pre-trained BERT model \cite{devlin-etal-2019-bert}. It is a deep language model that leverages Transformer architecture to learn the contextual relations between words in a textual description. During training, it also uses a word masking mechanism that masks 15\% of the words with a token and enables BERT to learn the robust embeddings. We considered text representations of 768 size. 
\\~
\noindent \textbf{Projection Head.} Once image and text representations of 2048 and 768 dimensions respectively are obtained, we projected them to the same subspace of size 256 through the projection head. The incorporation of contextual loss on top of contrastive loss requires the two non-aligned sub-spaces to be on the same embedding subspace. Therefore, we have used the Projection Head for these representations to project them onto the space.  We used image and text projection layers as $W_I$ and $W_T$ respectively, as shown in Fig.~\ref{fig:contrastive_contextual} with an output dimension d = 256.  We followed a change introduced by \cite{bachman2019learning} i.e., not to use the non-linear projection between the representation and the contrastive embedding space, instead use only a linear projection to map from each encoder’s representation to the multi-modal embedding space. 

\subsection{Loss Function} This section describes the loss function used in ContextCLIP. The total loss is calculated as:
\begin{equation}
    \mathcal{L} = \mathcal{L}_{CLIP} + \alpha \times \mathcal{L}_{CX}
\end{equation}
Here, $\alpha$ is a constant with value of 0.5. $\mathcal{L}_{CLIP}$ is the contrastive loss computed on image and text representations of size 2048 and 768, respectively, to bring similar image-text pairs together and dissimilar image-text pairs farther apart. $\mathcal{L}_{CX}$ is the contextual loss calculated through contextual similarity defined in Sec.~ \ref{sec:cs} and is applied at the individual features of contrastive aligned image-text pair. 
\subsubsection{\textbf{CLIP Loss ($\mathcal{L}_{CLIP}$)}:} 
Our work is most closely related to Contrtastive Language-Image pre-training (CLIP) \cite{radford2021learning}. While standard image models jointly train an image feature extractor and a linear classifier to predict some label, CLIP jointly trains an image encoder and a text encoder to predict the correct pairings of a batch of (image, text) training examples. An overview of our method is illustrated in Fig.~\ref{fig:contrastive_contextual}. At a high-level view, our method first converts each of the images and text to its equivalent representations u and v by passing them through pre-trained image and text encoders. Image representations (v) are of dimension 2048, and the text representations (u) are of dimension 768. Image and text representation are connected through two contrastive losses at the data point level. The first loss function is an image-to-text contrastive loss \cite{zhang2020contrastive} for the i$^{th}$ pair:

\begin{equation}
    \label{eqn:i2t}
    \ell_i^{(v \xrightarrow{} u)} = \log \frac{\exp{\langle \textbf{v}_i,\textbf{u}_i \rangle / \tau}}{\sum_{k=1} ^{N} \exp{\langle \textbf{v}_i,\textbf{u}_k \rangle / \tau}} 
\end{equation}
where, $\langle \textbf{v}_i,\textbf{u}_i \rangle$ represents the cosine similarity between image and text representations, i.e., $\langle \textbf{v},\textbf{u} \rangle = \textbf{v}^\top \textbf{u}/||\textbf{v}||||\textbf{u}||$; and $\tau \in \mathbb{R}^+$ represents a temperature parameter. This loss is InfoNCE \cite{oord2018representation} that tries to predict $\langle \textbf{v}_i,\textbf{u}_i \rangle$ as the true pairs and maximizes the mutual information between image and text representations. Hence, similar image-text pairs are put together, and the dissimilar image-text representations are put apart farther from each other. Here, contrastive loss is used between inputs of different modalities. Therefore, a similar text-to-image contrastive loss \cite{zhang2020contrastive} is defined as follows:
\begin{equation}
    \label{eqn:t2i}
    \ell_i^{(u \xrightarrow{} v)} = \log \frac{\exp{\langle \textbf{u}_i,\textbf{v}_i \rangle / \tau}}{\sum_{k=1} ^{N} \exp{\langle \textbf{u}_i,\textbf{v}_k \rangle / \tau}} 
\end{equation}

Thus, final training loss is a weighted average of image-to-text and text-to-image loss over all positive image-text pairs in each minibatch:

\begin{equation}
    \mathcal{L}_{CLIP} = \frac{1}{N} \sum_{i=1} ^{N} \left( \lambda \ell_i^{(v \xrightarrow{} u)} + (1- \lambda) \ell_i^{(u \xrightarrow{} v)} \right)
\end{equation}
where $\lambda \in \left[0,1\right]$ is a scalar weight. The maroon arrow in Fig.~\ref{fig:llda} shows the contrastive alignment between the image and text pair.

\subsubsection{\textbf{Contextual loss $\mathcal{L}_{CX}$}}
The connected representations learned through CLIP \cite{radford2021learning} loss are projected to 256-dimensional space, where they are brought closer together at the feature level. Contextual loss is the loss function targeted at non-aligned data. i.e., it does not require the two domains to be spatially aligned. It is based on both context and semantics and compares the regions of image and text with similar meanings while considering the context of the entire image. Our idea is to consider a batch of images and texts as a collection of features and measure the contextual similarity between image and text features based on the similarity between their features. \\~

\noindent \textit{Contextual Similarity:} \label{sec:cs}
We define a measure of similarity between a pair of text and image. To accomplish this, we represent each image $U$ encoded by ResNet \cite{he2016deep} and each text $V$ encoded by BERT \cite{devlin-etal-2019-bert} as a set of low-dimensional feature points in the same subspace via their projection heads $W_I$ and $W_T$, as shown in Fig.~\ref{fig:contrastive_contextual}. Feature points of image and text are represented as $U=\{u_i\}$ and $V=\{v_i\}$ as shown in Fig. ~\ref{fig:contrastive_contextual}. We assume $\lvert U \rvert = \lvert V \rvert =256$ in a batch of image-text pairs. To calculate the similarity between the images and texts, we find for each $v_j$, the feature $u_i$ that is most similar to it, and then sum the corresponding feature similarity values overall $v_j$. Formally, the contextual similarity $CX_{ij}$ between U and V is defined as follows.
\begin{equation}
CX(U,V)=\frac{1}{N}\sum_j\max_jCX_{ij}
\end{equation}

where, $CX_{ij}$ is the similarity between features $u_i$ and $v_j$.
 
The contextual similarity $CX_{ij}$ between points at the embedding space of image-text modalities helps for contextual cross-modal alignment by encouraging the distance between the image and text embedding to be close. Specifically, we consider feature $u_i$ as contextually similar to feature $v_j$, if it is significantly closer to it than to all other features in U. If this is not the case, i.e., $u_i$ is not more similar to any specific $v_j$, then its contextual similarity to all $y_j$ should be low. This method is resistant to distance scales, i.e., if $u_i$ is far from all $v_j$, then $CX_{ij}$ will be low for all $j$.\\~

\noindent \textit{On Image-Text latent space:} When the image and text pairs are similar, a one-to-one mapping exists between all the feature points. When they are dissimilar, a many-to-one mapping exists between text and image. A many-to-one mapping exists between the image and text pair, as shown in Fig.~\ref{fig:lldb}. This indicates the image-text pair are dissimilar in nature. We measure contextual similarity and aim to find a one-to-one mapping between image and text pairs to make them similar at the feature level to make them similar. The distance between image and text is a function of the near field. Contextual learning objectives use the nearest neighbor field in order to measure the similarity between images and texts. To avoid the need for geometric alignment between two modalities, it considers a particular feature of one modality and identifies the most similar features of the other modality. This yields the nearest neighbor field of matches between two modalities. The green arrow represents the nearest neighboring field. The loss function will use the green arrows and their associated weights. We start by computing a full affinity matrix between the point sets representing the images and texts, as shown in Fig.~\ref{fig: problem_defb}. Taking the maximum over each affinity yields the nearest neighbor field. To incorporate the context of each point, we use contextual affinity. We consider the context by normalizing the affinities in a softmax manner. The normalization is done in each row of the affinity matrix. Thus, we measure the similarity between two points representing an image and text using contextual affinity.

Matches between features are formed by considering all the features in the batch of images. This incorporates the cross-modal global image context into the similarity measure. The similarity between images and texts is then defined based on the similarity between the matched features. This allows the aligned image to spatially deform with respect to the text description. Let $d_{ij}$ represent the cosine similarity of $u_i$ and $v_j$. When $d_{ij} << d_{ik} $ and for all $k \neq j$, we consider features $u_i$ and $v_j$ to be similar. To begin, we normalize the distances, as done by \cite{mechrez2018contextual}:

\begin{equation}
    \tilde{d}_{ij} = \frac{d_{ij}}{\min_k d_{ik} + \epsilon}
\end{equation}
for a fixed $\epsilon=1e-5$. We shift from distances to similarities by exponentiation:
\begin{equation}
    w_{ij} = \exp{\left(\frac{1-\tilde{d_{ij}}}{h} \right)}
\end{equation}
where $h>0$ is a band-width parameter. Finally, the contextual similarity between features is defined to be a scale invariant version of the normalized similarities. In the extreme case, $CX(X, Y)$ is equivalent to counting how many features in V are the nearest neighbors of a feature in U, which is exactly the template matching measure proposed by \cite{talmi2017template}.
\begin{equation}\label{eq: cxij}
    CX_{ij} = w_{ij}/\sum_k w_{ik}
\end{equation}
Therefore, contextual loss between image-text space is defined as 
\begin{equation}
    \mathcal{L}_{CX}(u,v,d) = -\log \left( CX \left( R^d(u), B^d(v)
    \right) \right)
\end{equation}
The optimal solution to the contrastive loss formulation for CLIP would push the similarity between the normalized embeddings of the matched pairs towards one while forcing all other pairs of similarities to zero. Thus, aligning the cross-modal similarity matrix and minimizing the cross-modal alignement loss. However, this idealized scenario does not occur in practice, and we find that explicit alignment via contextual loss in ContextCLIP facilitates improved learning, as we show in our experiments. The pseudo-code for our approach is present in the supplementary material.
\begin{figure*}[!htb]
    \begin{minipage}{0.16\linewidth}\centering
	Text input. 
    \end{minipage}
    \begin{minipage}{0.41\linewidth}\centering
	(a) CLIP \cite{radford2021learning} 
    \end{minipage}
     \hfill
    \begin{minipage}{0.41\linewidth}\centering
	(b) ContextCLIP  (ours)
    \end{minipage}
    \begin{minipage}{0.16\linewidth}
    A boy in a red suit plays in the water
    \end{minipage}
    \begin{minipage}{0.41\linewidth}
        \includegraphics[width=0.98\linewidth]{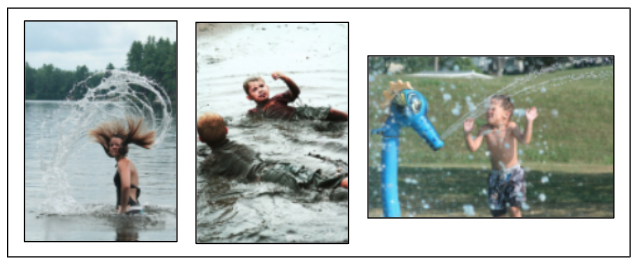}
    \end{minipage}
     \hfill
    \begin{minipage}{0.41\linewidth}
        \includegraphics[width=0.98\linewidth]{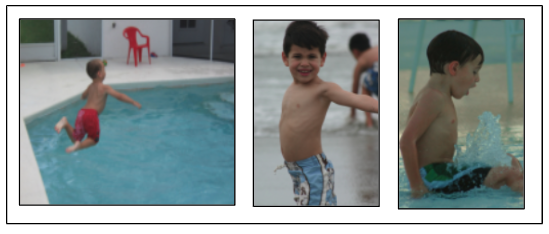}
    \end{minipage}
    \begin{minipage}{0.16\linewidth}
    Two girls enjoy a ride at an amusement park
    \end{minipage}
    \begin{minipage}{0.41\linewidth}
        \includegraphics[width=0.98\linewidth]{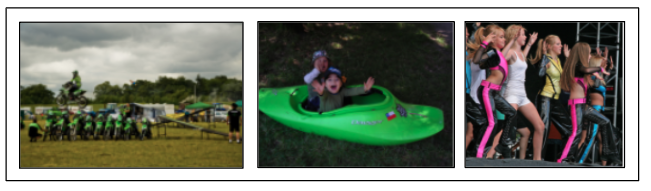}
    \end{minipage}
    \begin{minipage}{0.41\linewidth}
        \includegraphics[width=0.98\linewidth]{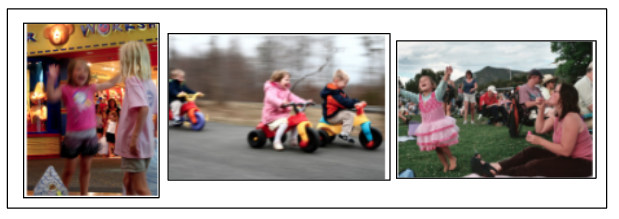}
    \end{minipage}   \begin{minipage}{0.16\linewidth}
   A person riding a skateboard jumps high above the concrete steps
    \end{minipage}
       \begin{minipage}{0.41\linewidth}
        \includegraphics[width=0.98\linewidth]{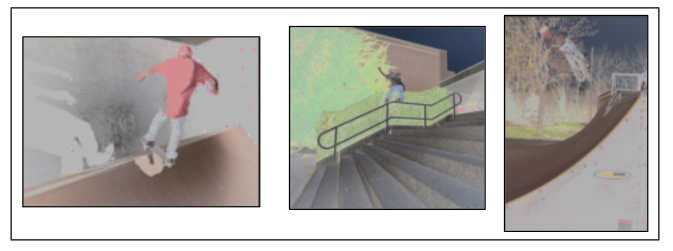}
    \end{minipage}
    \begin{minipage}{0.41\linewidth}
        \includegraphics[width=0.98\linewidth]{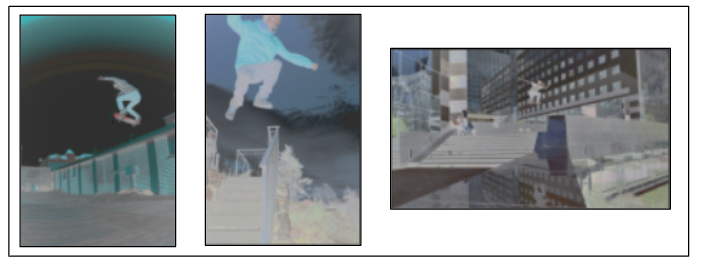}
    \end{minipage}       \begin{minipage}{0.16\linewidth}
	People hold flags in a crowded block
    \end{minipage}
    \begin{minipage}{0.41\linewidth}
         \includegraphics[width=0.98\linewidth]{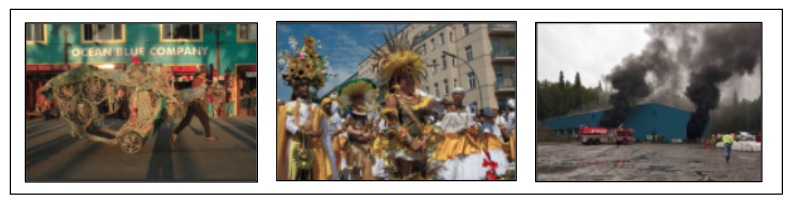}
    \end{minipage}
    \begin{minipage}{0.41\linewidth}
        \includegraphics[width=0.98\linewidth]{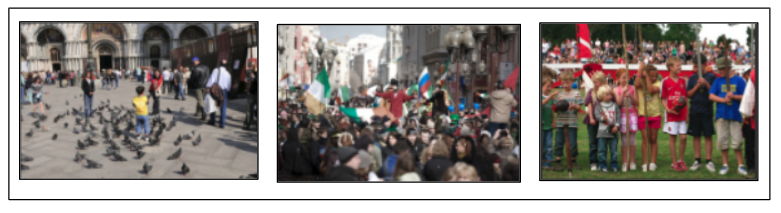}
    \end{minipage}
    \begin{minipage}{0.16\linewidth}
	A pale dog is running along a dirt path
    \end{minipage}
   \begin{minipage}{0.41\linewidth}
         \includegraphics[width=0.98\linewidth]{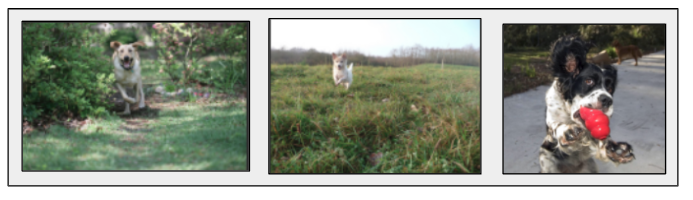}
    \end{minipage}
    \begin{minipage}{0.41\linewidth}
        \includegraphics[width=0.98\linewidth]{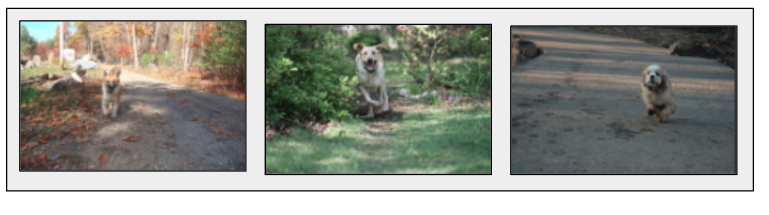}
    \end{minipage}
    \caption{Text-to-image Retrieval examples. (a) Represents the retrieval from CLIP \cite{radford2021learning} model. (b) Represents the retrieval from ContextCLIP model.}
    \label{fig:t2ir}
\end{figure*}

\section{Experimental Results}
We performed zero-shot transfer, fine-tuning, and text-to-image retrieval experiments on the CLIP and ContextCLIP methods. In a zero-shot transfer experiment, we observe that ContextCLIP improves top-1 accuracy over CLIP by 1 on CIFAR-10, by 0.5 on CIFAR-100, and by 2 on the BirdsNap dataset. ContextCLIP also outperforms CLIP in fine-tuning experiments, with an average relative gain of top-1 accuracy of +8 on CIFAR-10, +3 on RESISC45, and +1.5 on BirdsNap datasets. We also observed improved structure of embeddings on ContextXLIP in the representation space when projected to 2 dimensional space with t-SNE \cite{van2008visualizing} and discovered that representations with contextual loss are more structured than representations only with contrastive loss.
\subsection{Evaluation}
A zero-shot image classification task (Sec.~\ref{subsec:zs}) has been performed for both CLIP \cite{radford2021learning} and ContextCLIP, i.e., classifying test images into categories not seen at training time. We have used top-1 and top-5 accuracy to report the accuracy of predictions of classes given an image and a text template. We follow the evaluation strategy as suggested by \cite{radford2021learning} for zero-shot classification using prompt engineering. We used the following datasets for evaluating zero-shot transfer and fine-tuning experiments : CIFAR-10 \cite{krizhevsky2009learning}, CIFAR-100 \cite{krizhevsky2009learning}, MNIST \cite{deng2012mnist}, RESISC45 \cite{cheng2017remote}, BirdsNap \cite{berg2014birdsnap}. For each dataset, we use the names of classes to form a set of natural sentences such as 'a photo of the {class name}', 'a sketch of the {class name}' and more. Then, the similarity of the test image to each text template is computed (e.g., cosine distance), and the model predicts the category for which the image-caption similarity is highest. For evaluation of representations obtained, we performed fine-tuning experiments as mentioned in Sec.~\ref{subsec:ft}
\subsection{Zero-Shot Transfer} \label{subsec:zs}
For this experiment, we have trained our model on the Conceptual Captions \cite{sharma2018conceptual} dataset. The complete details of the dataset and experimental setup are mentioned in Sec.~\ref{sec:exp_set}. There is a pre-defined text template corresponding to each dataset as adopted from \cite{radford2021learning}. These templates are passed through text encoder to get a set of text embeddings for that class. To obtain a single text embedding for that class, this set of text embeddings is $ell_2$-normalized, averaged, and further $ell_2$-normalized. For a given image, the image embedding is obtained by passing it through an image encoder as defined in Sec.~\ref{sec:approach}. The class whose text embedding is closest to the test image is taken to be the predicted label. The top-1 and top-5 accuracies of zero-shot performance are presented in Table \ref{tab:zeroshot}.
\\~
CLIP \cite{radford2021learning} showed positive zero-shot transfer on CIFAR-10 and CIFAR-100 but negative zero-shot transfer on BirdsNap, RESISC45, and MNIST. Unlike CLIP, our method ContextCLIP showed positive zero-shot transfer performance on the BirdsNap (+1.91) and MNIST (+6.16) datasets in addition to CIFAR-10 (+1.03) and CIFAR-100 (+0.23) but negative zero-shot transfer performance on RESISC45 (-1.93). We reason that the positive zero-shot transfer of ContextCLIP on BirdNap and MNIST is because of the contextual loss in addition to the contrastive objective, which helps to align context (word) information from text description to the style information of the image. For RESISC45, ContextCLIP has shown negative zero transfer because of not being able to capture the alignment information from textual description and image. Among all of the evaluated datasets, MNIST has shown the most positive zero-shot transfer. We reason that this is because of the simple alignment relation between image and text space, as opposed to BirdsNap, CIFAR-10, and CIFAR-100, which have comparatively complex attributes and hence complex alignment relations. 

\begin{table*}[!h]
\renewcommand{\arraystretch}{1.3}
\caption{The table shows quantitative performance for Zero-shot Transfer on standard datasets.}
\begin{tabular}{|l|cc|cc|}
\hline
\multicolumn{1}{|c|}{\multirow{2}{*}{Dataset}} & \multicolumn{2}{c|}{\textbf{Top-1 Accuracy}}     & \multicolumn{2}{c|}{\textbf{Top-5 Accuracy}}     \\ \cline{2-5} 
\multicolumn{1}{|r|}{}                         & \multicolumn{1}{l|}{CLIP \cite{radford2021learning}}  & ContextCLIP (ours) & \multicolumn{1}{l|}{CLIP \cite{radford2021learning}} & ContextCLIP (ours) \\ \hline
\textbf{CIFAR-10} & \multicolumn{1}{l|}{9.37}     & \textbf{10.40}            & \multicolumn{1}{l|}{49.77}     & \textbf{50.14}            \\ \hline
 \textbf{CIFAR-100} & \multicolumn{1}{l|}{0.92}     & \textbf{1.15}            & \multicolumn{1}{l|}{4.34}     & \textbf{5.32}            \\ \hline
 \textbf{RESISC45} & \multicolumn{1}{l|}{3.41}     & 1.48            & \multicolumn{1}{l|}{12.19}     & 11.05            \\ \hline
 \textbf{BirdsNap} & \multicolumn{1}{l|}{8.5}     & \textbf{10.41}            & \multicolumn{1}{l|}{41.11}     & \textbf{46.86}            \\ \hline
 \textbf{MNIST} & \multicolumn{1}{l|}{12.26}     & \textbf{18.42}            & \multicolumn{1}{l|}{50.90}     & \textbf{59.82}            \\ \hline
\end{tabular}
\label{tab:zeroshot}
\end{table*}

\begin{table*}[!h]
\renewcommand{\arraystretch}{1.3}
\caption{The table shows quantitative performance for Fine-Tuning on standard datasets.}
\begin{tabular}{|l|cc|cc|}
\hline
\multicolumn{1}{|c|}{\multirow{2}{*}{Dataset}} & \multicolumn{2}{c|}{\textbf{Top-1 Accuracy}}     & \multicolumn{2}{c|}{\textbf{Top-5 Accuracy}}     \\ \cline{2-5} 
\multicolumn{1}{|r|}{}                         & \multicolumn{1}{l|}{CLIP \cite{radford2021learning}}  & ContextCLIP (ours) & \multicolumn{1}{l|}{CLIP \cite{radford2021learning}} & ContextCLIP (ours) \\ \hline
\textbf{CIFAR-10} & \multicolumn{1}{l|}{2.46}     & \textbf{10.40}            & \multicolumn{1}{l|}{60.78}     & \textbf{68.14}            \\ \hline
\textbf{CIFAR-100} & \multicolumn{1}{l|}{1.00}     & \textbf{1.22}            & \multicolumn{1}{l|}{5.37}     & \textbf{7.24}            \\ \hline
\textbf{RESISC45} & \multicolumn{1}{l|}{82.83}     & \textbf{85.38}            & \multicolumn{1}{l|}{98.30}     & \textbf{98.37}            \\ \hline
\textbf{BirdsNap} & \multicolumn{1}{l|}{14.25}     & \textbf{15.89}            & \multicolumn{1}{l|}{35.9}     & \textbf{43.85}            \\ \hline
\textbf{MNIST} & \multicolumn{1}{l|}{92.78}     & \textbf{94.47}            & \multicolumn{1}{l|}{99.12}     & \textbf{99.64}            \\ \hline
\end{tabular}
\label{tab:fine_tuning}
\end{table*}

\subsection{Fine-tuning} \label{subsec:ft}
The representation capabilities of our model ContextCLIP are evaluated through fine-tuning experiment. There are many ways to evaluate the quality of representations, e.g., One approach involves fitting a linear classifier on a representation extracted from the model and measuring its performance on various datasets. An alternative to this is measuring the performance of end-to-end fine-tuning the model. We adopted fine-tuning of the model because it increases the flexibility, and prior work has demonstrated that fine-tuning outperforms linear classification on most image classification datasets \cite{kornblith2019better}. We conducted fine-tuning experiments of the pre-trained representations obtained from the image encoder and text encoder i) only with contrastive loss and ii) contextual loss in addition to contrastive loss. The results are shown in the Table \ref{tab:fine_tuning}. We observe that top-1 classification accuracy improved on all datasets with the addition of contextual loss.
\\~
We observe that the maximum improvement in this experiment is obtained on CIFAR-10 (+7.94). Followed by this is RESISC45 (+2.55); CIFAR-100 (+.22); MNIST (+1.69); and BirdsNap (+1.64). CIFAR-10 showed maximum improvement as compared to other datasets because it has only 10 class categories. MNIST also has the same number of classes, but because of the very simple dataset and the complex formulation of the model, it is not able to learn well. Improvement in the evaluated datasets is shown in below order, i.e., CIFAR-10 RESISC45, which has 45 classes, followed by CIFAR-100 (45 classes), and BirdNap (200 classes). We observed that with the increase in the number of classes in datasets, the quality of representations learned from the supervision of other modalities reduces.

\subsection{Text-to-Image Retrieval}
For qualitative performance evaluation, we performed text-to-image retrieval tasks, i.e., given a text description, we retrieved a set of images with the CLIP and ContextCLIP methods, respectively. As shown in Fig.~ \ref{fig:t2ir}, the figures present on the left hand side are the images retrieved with the CLIP and on the right hand side are the images retrieved from ContextCLIP. We found that with the ContextCLIP method, the retrieved images are more semantically matched with the text description as compared to the CLIP method. e.g., corresponding to the text "A boy in red suit plays in the water," images retrieved with ContextCLIP contain either complete text information or most parts of text in the image. In the CLIP output, it does not show the boy in a red suit. In one of the retrievals, it shows two kids playing in water. Such specific number and color information is missing in the CLIP method, but ContextCLIP is capable of capturing such minute details of the text. Similarly, considering the other text examples like "People hold flags in a crowded block. Here, CLIP has not retrieved any images where either people or a flag can be seen in any image. As opposed to this, ContextCLIP has retrieved somewhat relevant results here. In some of the images, we can see people holding flags, a crowd somewhere, etc.

\section{Ablation Studies}
\begin{figure}[!htb]
    \begin{minipage}{0.25\linewidth}\centering
    Dataset
    \end{minipage}
    \begin{minipage}{0.35\linewidth}\centering
	(a) CLIP \cite{radford2021learning}   
    \end{minipage}
     \hfill
    \begin{minipage}{0.35\linewidth}\centering
	(b) ContextCLIP
    \end{minipage}
    \hfill
    \begin{minipage}{0.25\linewidth}\centering
	CIFAR-10
    \end{minipage}
    \begin{minipage}{0.35\linewidth}
        \includegraphics[width=0.98\linewidth]{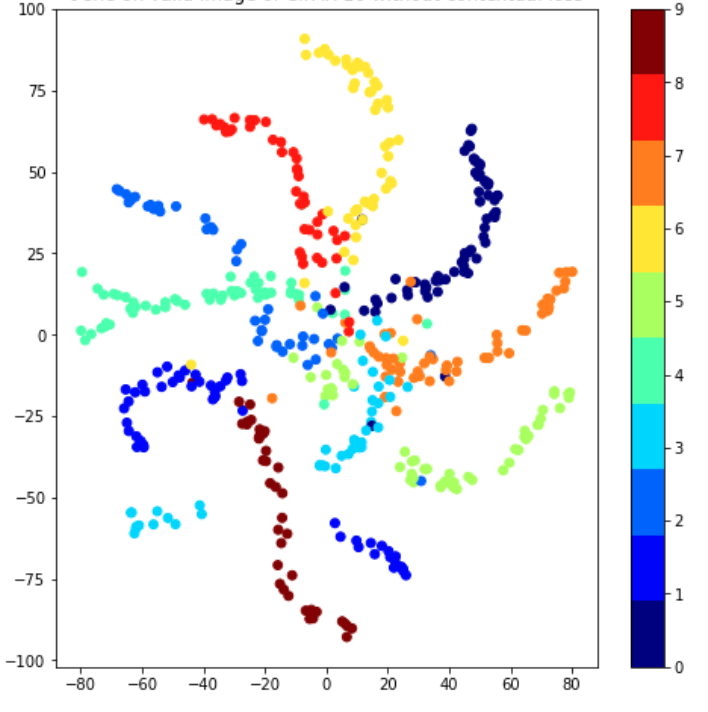}
    \end{minipage}
     \hfill
    \begin{minipage}{0.35\linewidth}
        \includegraphics[width=0.98\linewidth]{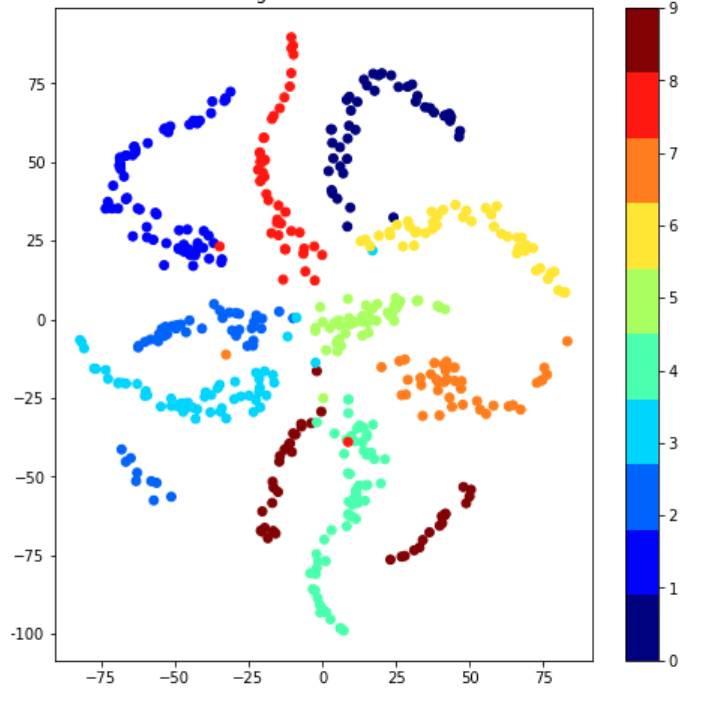}
    \end{minipage}
    \begin{minipage}{0.25\linewidth}\centering
	MNIST
    \end{minipage}
    \begin{minipage}{0.35\linewidth}
        \includegraphics[width=0.98\linewidth]{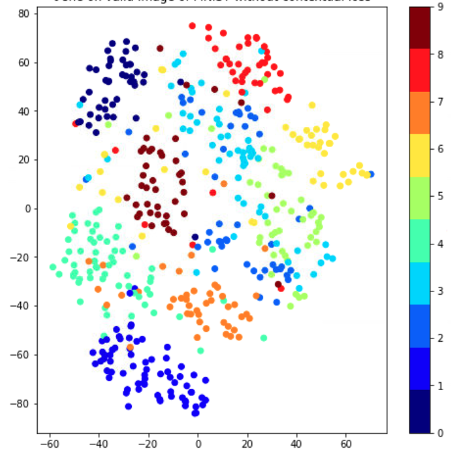}
    \end{minipage}
     \hfill
    \begin{minipage}{0.35\linewidth}
        \includegraphics[width=0.98\linewidth]{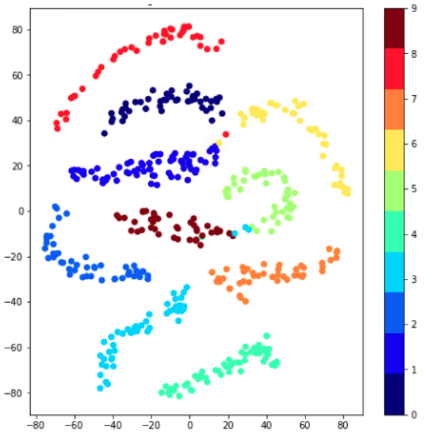}
    \end{minipage}
    \caption{Image Embeddings. The top and bottom rows are the embeddings of CIFAR-10 and MNIST, respectively. a) The image embeddings without contextual loss are shown in the left column; b) The image embeddings with contextual loss are shown in the right column.}
    \label{fig:ie}
\end{figure}

\begin{figure}[!htb]
    \begin{minipage}{0.25\linewidth}\centering
	Dataset
    \end{minipage}
    \begin{minipage}{0.35\linewidth}\centering
	a) CLIP \cite{radford2021learning}   
    \end{minipage}
     \hfill
    \begin{minipage}{0.35\linewidth}\centering
	b) ContextCLIP
    \end{minipage}
    \hfill
    \begin{minipage}{0.25\linewidth}\centering
	CIFAR-10
    \end{minipage}
    \begin{minipage}{0.35\linewidth}
        \includegraphics[width=0.98\linewidth]{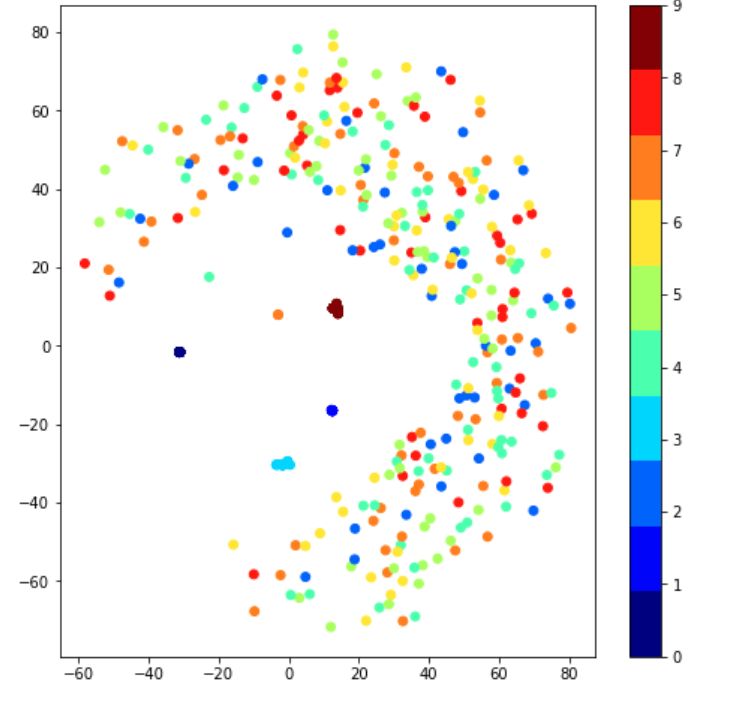}
    \end{minipage}
     \hfill
    \begin{minipage}{0.35\linewidth}
        \includegraphics[width=0.98\linewidth]{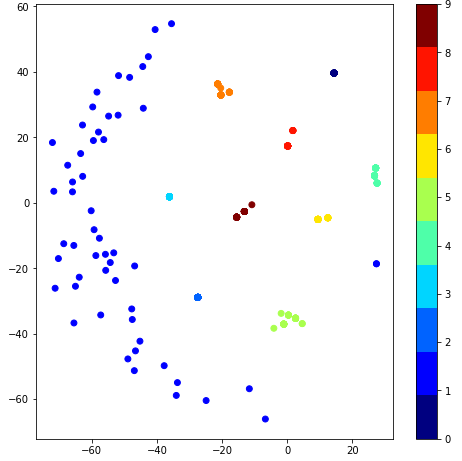}
    \end{minipage}
    \begin{minipage}{0.25\linewidth}\centering
	MNIST
    \end{minipage}
    \begin{minipage}{0.35\linewidth}
        \includegraphics[width=0.98\linewidth]{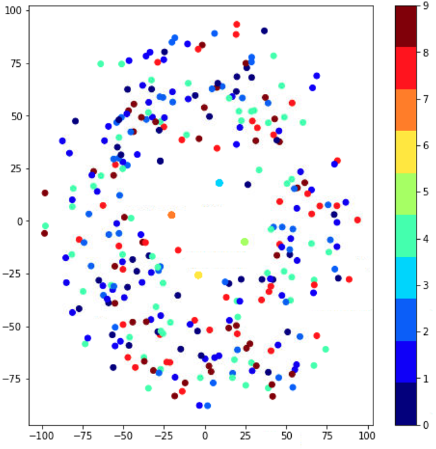}
    \end{minipage}
     \hfill
    \begin{minipage}{0.35\linewidth}
        \includegraphics[width=0.98\linewidth]{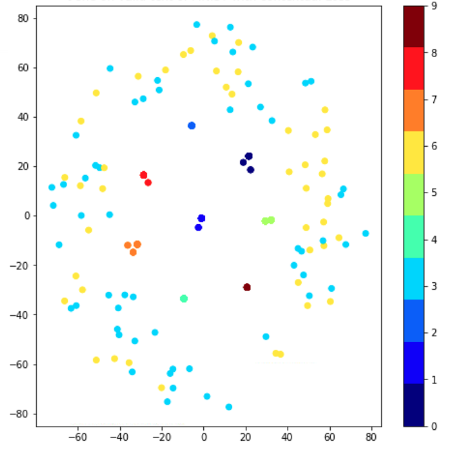}
    \end{minipage}
    \caption{Text Embeddings. The top and bottom rows are the embeddings of CIFAR-10 and MNIST, respectively. a) The left column shows text embeddings that have no contextual loss; b) the right column shows image embeddings that have contextual loss. }
    \label{fig:te}
\end{figure}

We projected high-dimension representations of image and text embeddings (256) obtained from their respective encoders (R and B) and followed by their projection heads ($W_I$ and $W_T$) to reduced-dimensional space. Low-dimensional projection is of size 2 and is done with t-SNE \cite{van2008visualizing} for CIFAR-10 and MNIST with and without contextual loss as shown in Fig.~\ref{fig:ie} and Fig.~ \ref{fig:te}. For both the datasets, image embeddings are more structured as compared to text embeddings. This is because, through ContextCLIP, we only aim to modify and learn robust image representations through natural language supervision. Once the image and text representations are obtained from the projection head, text representations do not get modified. With the addition of contextual loss, only image embeddings are updated with respect to the content and style of a given text description.


As shown in Fig.~\ref{fig:ie}, image embeddings clearly visualises all of the classes for both the datasets with the different colours. Image embedings of all classes are more clustered with ContextCLIP than CLIP. As opposed to image embeddings, there is no structured information in text embeddings \ref{fig:te}, neither in CLIP nor in ContextCLIP, but still comparatively better clustering information is available with ContextCLIP than CLIP. We tested text embeddings though BERT \cite{devlin-etal-2019-bert} as well as SimCSE\cite{gao-etal-2021-simcse}. In the figure \ref{fig:te}, we have shown the results with SimCSE. In SimCSE, text representations are learned in an unsupervised way.

\section{Experimental Setup} \label{sec:exp_set}
\noindent \textbf{Training Details.}
We have used \cite{openaiclipgithub} code for producing all tables and figures of our work. For contrastive learning, we used a temperature value of $\tau = 1.0$ and a loss weight of $\lambda = 0.75$. We used the Adam optimizer \cite{kingma2014adam} with a weight decay of 1e-3. A learning rate of 1e-4 and 1e-6 was used for the image encoder and text encoder respectively. We use a fixed batch size of 32. We train the model for 30 epochs. The network is trained jointly to minimise the distance between the modality-specific embeddings. \\~

\noindent \textbf{Datasets.} We used Conceptual Captions 1M \cite{sharma2018conceptual} (CC1M) image-caption pairs as the source of image and text pre-training data for zero-shot transfer experiments. Although this dataset is smaller than the custom dataset (400 million pairs) used in the original work on CLIP \cite{radford2021learning}, it is suitable for our available data and compute. Many subsequent works on language-image pre-training have used CC1M for benchmark evaluations\cite{li2021supervision, mu2021slip, Tejankar2021fistful, carlini2021poisoning}.  

\noindent \textbf{Fine-tuning:} For fine-tuning purposes, we trained variants of the ContextCLIP model end-to-end with the CIFAR-10 \cite{krizhevsky2009learning}, CIFAR-100 \cite{krizhevsky2009learning}, MNIST \cite{deng2012mnist}, RESISC45 \cite{cheng2017remote}, and BirdsNap \cite{berg2014birdsnap} datasets. We considered 80–20–10\% as the training, validation, and testing configurations. We considered the same datasets for evaluating the results of zero-shot transfer experiments. 
\section{Conclusion}
We presented ContextCLIP, a framework for contextual alignment of image-text pairs on CLIP visual representations. We studied the misalignment of image and text representations on CLIP representations. The main benefit of our method comes from incorporating contextual alignment into the joint image-text embedding space in addition to contrastive alignment. Empirically, we showed that ContextCLIP performs better than CLIP on zero-shot classification on datasets like
MNIST, CIFAR-10, CIFAR-100, BirdsNap and RESISC45 where CLIP had shown negative zero-shot transfer performance. 
For future work, we aim to evaluate our proposed method on ImageNet and its variants. We also showed that the representations learned by ContextCLIP are consistent with the concept-level knowledge of text description, as evidenced by text-to-image retrieval tasks. Also, with the fine-tuning experiments, we observed more robustness in visual representations than CLIP. We believe this work can motivate further studies on identifying conditions and regularization strategies under which the learned representations are synergistic across the image-text modalities for downstream applications. One significant future direction for ContextCLIP is to extend the work from text-to-image retrieval tasks to text-to-image style transfer and manipulation tasks.

\bibliographystyle{ACM-Reference-Format}
\bibliography{ICVGIP-Latex-Template.bib}

\newpage

\lstset{%
  basicstyle=\small\ttfamily,
  frame=single,
  morecomment=[f][\color{blue}][0]{*},
  morecomment=[f][\color{red}][0]{\#},
  }


\titlenote{Supplementary Material}
\begin{figure*}[!ht]
\Large{\textbf{ContextCLIP: Contextual Alignment of Image-Text pairs on CLIP visual representations \vspace*{3pt}\\(Supplementary Material)}}\\~
\begin{lstlisting}[mathescape]
# Pseudo-code for ContextCLIP
1. $encoder_{image}$      - ResNet Model
2. $encoder_{text}$       - BERT Model 
3. $I[n, h, w, c]$      - minibatch of images 
4. $T[n, l]$            - minibatch of texts 
5. $W_I [d_i, d_e]$        - learned projection of image to embed 
6. $W_T [d_t, d_e]$        - learned projection of text to embed 
7. $t$                       - learned temperature parameter 
# extract feature representations of each modality 
8. $I_f$ = $encoder_{image}(I)$     - $[n, d_i]$ 
9. $T_f$ = $encoder_{text}(T)$      - $[n, d_t]$ 

# joint multimodal embedding [n, d_e] 
10. $I_e = l_2 normalize(np.dot(I_f, W_I), axis=1)$ 
11. $T_e = l_2normalize(np.dot(T_f, W_T), axis=1)$ 

# scaled pairwise cosine similarities [n, n] 
12. logits = $np.dot(I_e, T_e.T) * np.exp(t)$ 

# contrastive loss function 
13. labels = np.arange(n) 
14. $loss_i$= crossEntroLoss(logits, labels, axis=0) 
15. $loss_t$ = crossEntroLoss(logits, labels, axis=1) 
16. $loss_{contrastive}$ = $(loss_i + loss_t)/2$ 

# contextual loss function 
17. $loss_{contextual}$ = contextualLoss$(I_e, T_e)$

18. loss = $loss_{contrastive} + lambda \times loss_{contextual}$ 
\end{lstlisting}
    \caption*{Algorithm 1. This describes sequence of steps taken to add contextual loss on CLIP joint image-text embedding space.\\~}
    \label{fig:my_label}
\RaggedRight This details of Algorithm 1 for the proposed ContextCLIP is described as follows.
\begin{itemize}
\item Line 1 and 2 are the ResNet image encoder and BERT text encoder. 
\item Line 3 and 4 are the mini-batch of size n of images (I) and texts (T). 
\item Line 5 and 6 are the projection heads for image ($W_I$) and text ($W_T$) which projects the different size image ($d_i$) and text ($d_t$) representations to the same size ($d_e$). 
\item Line 7 is the temperature value used for calculating the contrastive loss between image and text pairs. 
\item Line 8 and 9 are the image ($I_f$) and text ($T_f$) representations obtained from the image-encoder and text-encoder respectively.  
\item Line 10 and 11 are the normalised image ($I_e$) and text ($T_e$) embeddings of their representations obtained by dot product of image and text representations with their projection heads.
\item Line 12 calculates the logits of image and text embeddings i.e. it calculates the image and text similarity with a specific temparture value. 
\item Line 13 defines the true labels of all images. 
\item Line 14 and 15 calculates the cross entropy loss $loss_i$ and $loss_t$ between image  and text  embeddings based on true and predicted image-text pairs. 
\item Line 16 calculates the contrastive loss between image and text embeddings. 
\item Line 17 calculates the contextual loss at the semantic level of image and text embeddings. 
\item Line 18 calculates the sum of contrastive and contextual loss. This completes the total loss calculation for a mini-batch of texts and images pair with the weightage of $\lambda = [0-1]$.
\end{itemize}

The same sequence of steps is performed for another batch of images and so on. This code was iterated for 25 epochs on training datasets: Conceptual Captions for zero-shot transfer experiments; CIFAR-10, CIFAR-100, MNIST, BirdsNap, RESISC45 for fine-tuning experiments; Flickr-8K for text-to-image retrieval experiments.  
\end{figure*}

\end{document}